\documentclass[10pt,emptycopyrightspace]{ewsn-proc}
\pdfoutput=1

\usepackage{hyperref}
\usepackage{comment}
\usepackage{xspace}
\usepackage{amsmath}
\usepackage{amssymb}
\usepackage{balance}
\usepackage[labelformat=simple]{subcaption}
\usepackage{booktabs} 
\usepackage{tabu}
\usepackage[flushleft]{threeparttable}
\usepackage{multirow}
\usepackage{array}
\usepackage[numbers,sort&compress]{natbib}
\usepackage{mathtools}
\usepackage{textcomp}
\usepackage{xcolor}
\usepackage{graphicx}
\graphicspath{{./figs/}}

\AtBeginDocument{%
	\providecommand\BibTeX{{%
			\normalfont B\kern-0.5em{\scshape i\kern-0.25em b}\kern-0.8em\TeX}}}

\newtheorem{theorem}{Theorem}
\newtheorem{definition}{Definition}
\newtheorem{lemma}{Lemma}
\newtheorem{corollary}{Corollary}
\newtheorem{remark}{Remark}

\newcommand{\prob}[1]{\Pr\left[#1\right]}
\newcommand{\optdistnotation}{\mathrm{MED}}
\newcommand{\optdist}[2]{\optdistnotation(#2 \vert #1)}
\newcommand{\region}[2]{\mathcal{X}(#1,#2)}
\newcommand{\crossentnotation}{\mathrm{CE}}
\newcommand{\crossent}[2]{\crossentnotation(#1,#2)}

\def\D{\mathcal{D}}
\def\G{\mathcal{G}}
\newcommand{\DZ}{\mathcal{D}_z} 
\newcommand{\DY}{\mathcal{D}_y}

\DeclareMathOperator*{\argmax}{arg\,max}

\newcommand{\denselist}{\itemsep 0pt\parsep=1pt\partopsep 0pt}
\newcommand{\bitem}{\begin{itemize}\denselist}
\newcommand{\eitem}{\end{itemize}}
\newcommand{\benum}{\begin{enumerate}\denselist}
\newcommand{\eenum}{\end{enumerate}}

\newcommand{\data}{\text{data}}

\begin{document} 
%



\numberofauthors{1}

\renewcommand\alignauthor{%
  \end{tabular}%
  \hspace{0pt}
  \begin{tabular}[t]{p{\auwidth}}\centering
}

\author{
\alignauthor Aria Rezaei$^{\ast}$, Chaowei Xiao$^{\dagger}$, Jie Gao$^{\ddagger}$, Bo Li$^{\circ}$, Sirajum Munir$^{\diamond}$ \\
    \affaddr{$^{\ast}$Stony Brook University, arezaei@cs.stonybrook.edu. 
    $^{\dagger}$NVIDIA, chaoweix@nvidia.com.
$^{\ddagger}$Rutgers University, jg1555@rutgers.edu.
$^{\circ}$UIUC, lbo@illinois.edu.
$^{\diamond}$Bosch Research, sirajum.munir@us.bosch.com}
}

\title{Application-driven Privacy-preserving Data Publishing with Correlated Attributes}

\maketitle

\begin{abstract}
Recent advances in computing have allowed for the possibility to collect large amounts of data on personal activities and private living spaces. To address the privacy concerns of users in this environment, we propose a novel framework called PR-GAN that offers privacy-preserving mechanism using generative adversarial networks. Given a target application, PR-GAN automatically modifies the data to hide sensitive attributes -- which may be hidden and can be inferred by machine learning algorithms -- while preserving the data utility in the target application. Unlike prior works, the public's possible knowledge of the correlation between the target application and sensitive attributes is built into our modeling.
We formulate our problem as an optimization problem, show that an optimal solution exists and use generative adversarial networks (GAN) to create perturbations. We further show that our method provides privacy guarantees under the Pufferfish framework, an elegant generalization of the differential privacy that allows for the modeling of prior knowledge on data and correlations. Through experiments, we show that our method outperforms conventional methods in effectively hiding the sensitive attributes while guaranteeing high performance in the target application, for both \emph{property inference} and \emph{training} purposes. Finally, we demonstrate through further experiments that once our model learns a privacy-preserving task, such as hiding subjects' identity, on a group of individuals, it can perform the same task on a separate group with minimal performance drops.
\end{abstract}
\section{Introduction}\label{sec:intro}
\begin{figure}[!t]
    \centering
    \includegraphics[width=0.99\columnwidth]{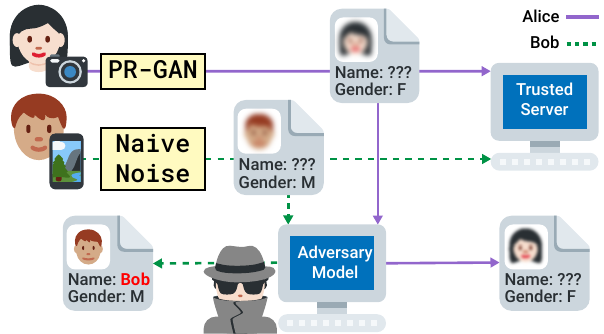}
    \caption{Alice and Bob send data from their personal devices to a trusted server, masking their identity while making their gender visible. An adversary cannot decipher Alice's identity as she has used our method (PR-GAN) while he can do so for Bob, who use a naive noise generator.
    }
    \label{fig:attack}
\vspace{-10pt} \vspace*{-4mm}
\end{figure}
Recent years have witnessed an explosive growth in ubiquitous sensing and 
data-driven AI innovations in every aspect of our lives. 
While we celebrate the convenience brought to us by these technologies, the collected data can often be personal information and contain attributes that could be extremely sensitive. 
For example, using background noise, one can infer the number of people talking, their identity
or the type of environment~\cite{Xu2013-ds,Miluzzo2008-ru,Lu2010-ft}.
Data from electricity meters can reveal sensitive information about the household
~\cite{Zeifman2014-gq}.
GPS trajectories can reveal social ties and are very unique for each person~\cite{unique_in_the_crowd,Wang2011-sy,Eagle2009-xq}.
Using WiFi scans, one can discover occupancy patterns of private households and localize individuals~\cite{conf/sensys/WifiActivityPattern,WiFiDataset}. 
This means that it is crucial that we find the correct balance between using data to improve the quality of our lives while making sure sensitive information is effectively protected in IoT systems.


There are numerous metrics for data privacy,
such as $k$-anonymity \cite{Sweeney:98:achieving}, differential privacy (DP)~\cite{Dwork06differential}, and its extension that supports correlations and prior knowledge on the data, namely Pufferfish privacy~\cite{kifer2014puffer} among others. By introducing noises either in the original data (for data publication) or the query output (for answering interactive queries on private data), these criteria provide guarantees that an attacker cannot infer substantially more with or without access to a particular record.
Most conventional methods are designed for structured datasets, like patient's medical records, with an \emph{a priori} knowledge of which one of the features in the data are deemed sensitive.
In this work, however, we focus on \emph{unstructured} data in sensor signals, such as images, binary vectors and other raw physical readings, where sensitive features are patterns hiding in the data that could be learned by an adversary.

Our focus in this work is to develop a novel framework that supports \emph{private data publication},
which is suited for distributed sensing and IoT settings. This is different from another common setting where the data is kept in a private database and noise is added to the results of a query to ensure privacy. A major advantage of our approach is that after publication, the data could be used for a wide variety of applications. Note also that 
if an entity publishes \emph{only} the target attributes, while ensuring a high degree of privacy, the published information could not have any further use; this is why this naive approach is usually not considered. 


Consider a use case in a smart building application, where the WiFi signal strength of occupants is routinely recorded to monitor the integrity of the local network and detect system failures.   The occupants have no problem contributing their data for such purposes as long as they don’t feel that their identities might be revealed or that their movements are tracked. To ensure this, the administrators propose a compromise:  the collected data might be accurate enough to reveal which \emph{floor} a person is on, but will not reveal which specific room they are in at any moment.  Note that here, inferring which floor a person is on is not the goal of publishing the data;  it just ensures an acceptable level of accuracy in the data.   Note  also  that  an  adversary  has partial information about the target and sensitive attributes: \emph{if a person is on the 3rd floor,  she could only be in one of the rooms on the 3rd floor}.  This information comes by definition and requires no extra effort to retrieve it.  We explicitly model this correlation, and by doing so, we are in effect defending against a stronger adversary compared to the prior work~\cite{dysan,olympus,minimax}.

Since the organization that keeps the data may not be trusted to properly address the privacy concerns,
a more preferred approach is ensuring privacy from the source. Figure~\ref{fig:attack} shows such a case where Alice and Bob want to share their photos with an external server and they use perturbations to conceal their identity. An adversary, using an auxiliary dataset has trained an identity-revealing model. She then intercepts the messages midway and manages to reveal the identity of Bob using this model. Our goal is that if Alice's photo, perturbed by our method, is intercepted, the adversary won't succeed in revealing her identity.


\def\x{\mathrm{x}}
\smallskip\noindent\textbf{Problem Definition and Our Contribution.}
We assume that the sensor data $\x$ is used to learn a \emph{target attribute} $f_y(\x)$. Meanwhile, there are a number of \emph{sensitive attributes} $f_z(\x)$ that are of concern in the sensor data. The function $f_y$ describes data utility and function $f_z$ characterizes the privacy concern. We assume that both the target attributes and the sensitive attributes are deeply embedded in the physical sensor signals. Even with anonymity, both attributes can be learned with supervised learning. Therefore,
the data collected from the individual sensors could easily reveal the sensitive attributes about the users who contribute the data.

We plan to address the problem by developing an algorithm $A$ that perturbs the raw sensor reading $\x$ to $\x'=A(\x)$ such that the utility of the system is high, as defined by the success of learning the target attribute, i.e., $f_y(\x')\approx f_y(\x)$, and the privacy of the sensitive attributes is protected, as measured by the significantly reduced performance of $f_z(\x')$ accurately predicting $f_z(\x)$.

We formulate the problem as an optimization problem:
we aim to minimize the performance loss of $f_y$ when perturbed data is used, and maximize the uncertainty among the predicted values of $f_z$ while adhering to the constraints of publicly known relation between $f_y$ and $f_z$.
We show the existence of an optimal perturbation function and analyze the privacy guarantees provided by our mechanism. 

To produce perturbations tailored to each data point, we use Generative Adversarial Networks (GAN)~\cite{goodfellow2014generative}.
A standard GAN is composed of two neural networks contesting with each other in a zero-sum game.
Two models are alternatively trained: a generative model $\G$ tries to produce \emph{fake} data records. The discriminator $\D$ tries to predict whether an input data is \emph{fake} (produced by $\G$) as accurately as possible, while $\G$ tries to maximize the probability of $\D$ making a mistake by producing artificial data that is indistinguishable from the real data. The GAN structure ensures that
the resulting data looks like the original data.
In our setting, we use two classifiers as approximations of  $f_y$ and $f_z$.
Through rigorous experiments, we show compelling evidence of the superior performance of our perturbations, compared to baseline approaches. 
Our key contributions are listed below:
\benum
\item We are, to the best of our knowledge, the first to explicitly model the correlation between target and sensitive attributes, which reflects real-world scenarios more closely. By doing so, we are assuming a stronger adversary, compared to prior work, who utilizes this information to mount a more successful attack.
\item Our models are light-weight and do not require the presence of the whole dataset to generate perturbations. Once trained on a portion of data, the noise generator can be deployed on mobile devices. In Section~\ref{sec:mobile} we compare the computational cost of our models with state-of-the-art models designed specifically to run on mobile devices.
\item In Section~\ref{sec:privacy_guarantee}, we establish a connection between maximizing an adversary's uncertainty about the sensitive labels and providing guarantees based on Pufferfish privacy, an extension of DP. The reason that this is our framework of choice is that it allows us to model correlations and prior knowledge on the data.
\item Through experiments, we show that once our model is trained to hide identity-revealing features from depth images of a group of individuals, it can perform the same task on a \emph{completely different} group of individuals with minimal performance loss. We discuss the benefits of this in Section~\ref{sec:transfer}.
\item Unlike prior work, we test the utility of the perturbed data on a \emph{training task} as well, showing that a well-performing classifier can be trained on our perturbed data. In Section~\ref{sec:new_training} we discuss this in more depth.
\eenum
Our framework is flexible and any model that supports gradient updates could be used instead of the ones used here. Additionally, unlike prior work, we don't transform the data into an entirely different feature space to protect privacy~\cite{KDTreePPDP,RotationPPDP}.
\section{Related Work}\label{sec:related_work}
\noindent\textbf{Privacy in Learning Algorithms}
A large body of work have focused on manipulating the existing training and inference algorithms to protect privacy of training data. For example, a differentially private training algorithm for deep neural networks in which noise is added to the gradient during each iteration~\cite{Abadi2016-lb,Phan2016-ub,Shokri2015-nj}, and a ``teacher-student" model using aggregated information instead of the original signals~\cite{papernot2016semi}. It is also proposed to train a classifier in the cloud, with multiple users uploading their perturbed data to a central node without ever revealing their original private data~\cite{CollaborativeLearningCloud,federated_learning}.
\noindent\textbf{Privacy-Preserving Data Publishing}
A different approach to preserve privacy is making sure sensitive elements of the data is removed before publishing it, often called privacy-preserving data publishing (PPDP)~\cite{bowden1992privacy}.
A main line of work focuses on transforming numerical data into a secondary feature space, such that certain statistical properties are preserved and data mining tasks can be done with minimal performance loss~\cite{KDTreePPDP,RotationPPDP,ProjectionPPDP}.
These methods guarantee data utility only on this secondary space. This means that a classifier trained on the perturbed data is not guaranteed to perform well on original data. This can be troublesome in a scenario where a classification model is trained on public, perturbed data and is going to be deployed on users' personal devices to perform a task locally on non-perturbed private user data. Our perturbed data can be used in conjunction with original data in specified target applications. In addition, some of the methods in this category rely on expensive computations which renders them infeasible on large datasets.

\noindent\textbf{Adversarial Learning}
GANs have been successfully used to produce \emph{adversarial examples} that can fool a classifier to predict wrong classes~\cite{xiao2018generating}. Some have formulated the problem of privacy protection as producing adversarial examples for an identity-revealing classifier~\cite{adversarial_face_privacy,context_aware_adversarial_privacy}. However, we demonstrate through experiments that the absence of a predictor function, $f_y$, that maintains the utility of the data, leads to a weaker utility guarantee for the published data.

\noindent\textbf{Adversarial Privacy}
There has been a recent surge in the use of adversarial learning for privacy protection. In this context, privacy is defined in various ways. Some have employed proxies for privacy such as reducing the mutual information between the data and sensitive attributes that is discolsed by publishing a perturbed dataset~\cite{information_theory_privacy}, or minimizing the distance between data records belonging to different \emph{sensitive} categories so as to make it harder to distinguish between them~\cite{maximize_within_class_sensitive,shredder}. The problem with employing such heuristics is that they do not directly aim to prevent an adversary from inferring sensitive information. Acknowledging this, others have tried to minimize an adversary's success in predicting sensitive information~\cite{dysan,olympus,minimax}. However, no connection between such an approach and rigorous privacy formulations have been established yet.
We are, to the best of our knowledge, the first to attempt to link the inability of an adversary to predict sensitive information to a known privacy framework, namely Pufferfish.
Another key difference between the prior work and ours is the assumption that the correlation between the target attribute, $y$, and the sensitive, $z$, is at least partially known.
As we show later on, the possession of such knowledge by an adversary is very likely in real-world scenarios.
This correlation is modeled directly into our formulation, which is unlike prior work where $y$ and $z$ are assumed to be uncorrelated~\cite{dysan,olympus}.
Some prior work rely on the availability of an \emph{acceptable} and an \emph{unacceptable} set of data records for public disclosure;
the model then tries to learn a transformation from the latter to the former~\cite{censorship_fairness}.
In real-world scenarios, this \emph{a priori} knowledge of what safe-to-publish data looks like is not always possible, and our model relies solely on the two classifiers, $f_y$ and $f_z$, to learn how the published results should look like. Finally, another line of work tries to create a private intermediate encoding of the data for specific neural network architectures~\cite{shredder}. This approach will make the use of the published dataset limited as it can only be used for neural networks with a certain architecture. In our work, we assume that the published data looks as much to the original data as possible and is ready-to-use once disclosed with the public.

\begin{figure*}[!tbh]
    \centering
    \includegraphics[width=0.95\textwidth]{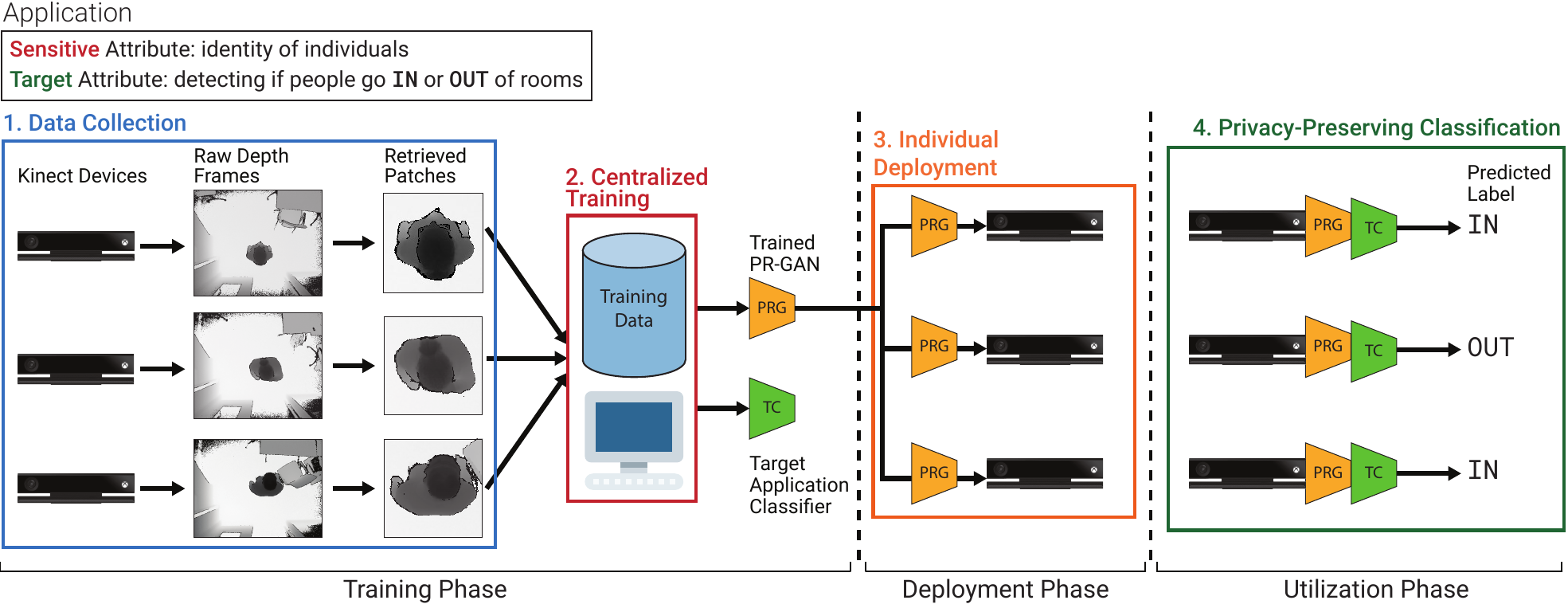}
    \caption{System overview for a smart building setting with specific target and sensitive attributes. In the \emph{training} phase, the manufacturer collects the data, labels it, and trains the noise generator (PRG) and the target classifier (TC). Then, in the \emph{deployment} phase, PRG modules are installed on individual sensors (Kinect devices). Finally, during the \emph{utilization} phase, images are first perturbed using PRG and then classified in a privacy-preserving manner with TC.
    More details in Section~\ref{section:system}.}    \label{fig:system_overview}
\end{figure*}
\section{System Overview}\label{section:system}
To demonstrate our system's capability in protecting privacy in a sensing system, we provide the following overview of a potential scenario.

Consider a smart building setting where the building managers would like to know the number of people in each room at any moment to maximize energy efficiency. To do this, they install sensing devices, such as Xbox Kinect devices, near entrances to each room to maintain an accurate count.
The occupants in the building, on the other hand, might be worried about being tracked by the system as their identities might be revealed using the collected data.

Figure~\ref{fig:system_overview} shows the overview of a privacy preserving sensing system. There are $3$ phases in total. During the \emph{training} phase, the \emph{manufacturer} of the noise-generating model, PR-GAN, mounts XBOX Kinect devices on top of entrance doors, similar to the setting in the target building, and asks volunteers to move under the device, producing instances with them facing ``inside'' or ``outside''. Here, the sensitive attribute is the subjects' identity while the target attribute is their facing direction. Then, the two classifiers on the target and sensitive attributes,
together with the noise-generating module, PR-GAN is trained.
Once training is over, during the \emph{deployment} phase, PR-GAN modules can be installed on individual sensors by either the manufacturers or the building managers. As we mentioned before, PR-GAN is light-weight and computationally inexpensive, hence suitable to run on mobile devices (more on this in Section~\ref{sec:mobile}).

Finally, in the \emph{utilization} phase, the collected imagery is first anonymized in real-time using the noise generator, and the noisy images are then fed to the target classifier for
privacy-preserving
classification.

\noindent\textbf{Threat Model}
We make the following assumptions about the adversary: (1) the adversary might get access to the noisy data, either in the event of a security breach or via other means. (2) The adversary could also possess a classifier on sensitive attributes by training a model on an auxiliary dataset. (3) Since we intend to publish the target labels, it is safe to assume that the adversary has access to them. (4) Partial information about the correlation between target and sensitive attributes is known by public, and hence accessible by the adversary.

\section{Theory}\label{sec:definition}
\subsection{Problem Definition and Optimality}

Suppose that we have a dataset $X$ with $n$ entries, where each entry is a vector $\x_i \in \mathbb{R}^N$ coming from an unknown distribution $\mathcal{P}_{\text{data}}$. Additionally, each entry is accompanied with a \emph{target} label, $y_i$, and a \emph{sensitive} label, $z_i$, where $y_i \in \mathcal{Y}$ and $z_i \in \mathcal{Z}$ are categorical features.

It is possible for an adversary, or the public, to have at least partial information about the relation between $y$ and $z$. We model this assumption by assuming that a conditional distribution of $z$ for a particular value of $y$,
denoted by $\theta_y$ where $\theta_y(z) = P(z \vert y)$, is known for a subset of $\mathcal{Z}_y \subseteq \mathcal{Z}$. The set of all such distributions of $\theta_y$ is denoted by $\Theta$.
Note that $\mathcal{Z}_y$ can be empty, representing the case where no prior knowledge is available.
As we will see later on in many scenarios, simply based on contextual definitions of $y$ and $z$, it is possible for a partial knowledge of $P(z|y)$ to be held by the public and we should expect that an attacker might exploit it to mount a more successful attack.

For instance, suppose that a dataset consists of images of $20$ subjects, $10$ of them male and the other $10$ female. Let the identity of the subjects be $z$ and their gender $y$. Since we intend to publish $y$, we should assume that once the data is published the gender of each subject is known to all (e.g.: $y = \text{male}$). Knowing this, an adversary can cross off the $10$ female subjects as potential $z$.
Since each identity can be mapped only to one gender (for the sake of simplicity), we can assume that two disjoint sets, $\mathcal{Z}_\text{male}$ and $\mathcal{Z}_\text{female}$, comprise all identities. Then, the prior knowledge about the labels would be $\theta_\text{male}(z) = 0 \: (z \in \mathcal{Z}_\text{female})$ and vice versa.

Finally, we have two functions $f_y \colon \mathbb{R}^N \mapsto \mathcal{Y}$ and $f_z \colon \mathbb{R}^N \mapsto \mathcal{Z}$, that predict $y$ and $z$ respectively.
The goal is to find a transition function, $g \colon \mathbb{R}^N \mapsto \mathbb{R}^N$,
such that one could still infer the target labels from $\x' = g(\x)$ but would fail to do so for sensitive labels with any certainty. In particular, we want to produce a perturbed dataset $X' = \{\x'=g(\x), \x\in X\}$ such that the following is minimized:
\begin{equation}\label{eq:obj_constraint}
\begin{aligned}
    &{} \mathbb{E}_{\x' \in X'}\left[ \ell(f_y(\x'), y) - H(f_z(\x') \vert y)\right],\\
    \text{s.t.},\, & \forall \theta_y \in \Theta, z \in \mathcal{Z}_y:
    \prob{f_z(\x') = z \vert y} = \theta_y(z).
\end{aligned}
\end{equation}
In the above, $y=f_y(\x)$, $\ell$ is a suitable loss function and $H(a \vert b)$ is the entropy of $a$ conditioned on knowing $b$. The first term of the objective function keeps target labels unchanged by $g(\x)$ while the second term introduces maximum uncertainty for predicted $z$ labels, $f_z(\x')$.
$X'$ is finally released for public use. The reason for the desirability of maximum uncertainty for $P(z|y)$ is to make sure that an adversary cannot learn from the published noisy images and the target labels anything that she already does not know.

\smallskip\noindent\textbf{Maximum Entropy Distribution}: To understand the objective function better, consider data set $X^*_z$ which maximizes the second term in \eqref{eq:obj_constraint}:
\begin{equation}\label{eq:max_cond_entropy}
\begin{aligned}
    &{} X^*_z = \argmax_{X'} \mathbb{E}_{\x' \sim X'}\left[ H(f_z(\x') \vert y \right]\\
    \text{s.t.},\, & \forall \theta_y \in \Theta, z \in \mathcal{Z}_y:
    \prob{f_z(\x') = z \vert y} = \theta_y(z).
\end{aligned}
\end{equation}
The conditional distribution for $z$ with respect to $y$ which maximizes the above is known as
the \emph{maximum entropy distribution} (MED) for predicted $z$ values of $X^*_z$ given $\Theta$:
\begin{equation}\label{eq:max_entropy_dist}
\optdist{y}{z} =
    \begin{cases}
    \theta_y(z)&  \text{if } z \in \mathcal{Z}_y \\
    {c_y}/k& \text{otherwise}
    \end{cases},
\end{equation}
where $c_y = 1 - \sum_{z \in \mathcal{Z}_y} \theta_y(z)$ and
$k = \left|\mathcal{Z} \setminus \mathcal{Z}_y \right|$. 

Intuitively, to prevent the attacker from learning anything she does not already know (via prior, public knowledge $\Theta$), $\optdistnotation$ takes the form of a uniform distribution in the absence of any prior knowledge $\theta_y$.
Notice that $\optdistnotation$ depends \emph{not} on $\x$, but solely on each data point's target attribute $y$, which is assumed to be publicly known.

To understand $\optdistnotation$ better, consider the face image dataset again. Imagine that the target labels are ``male'' and ``female'' and there are only $4$ female subjects: $f_1$, $f_2$, $f_3$ and $f_4$. Without any additional information, given that an anonymized image has label ``female'', the resulting  $\optdistnotation$ will assign $1/4$ probability of that image belonging to any of the $4$ female subjects. However, if the adversary knows that the image is selected at random and half of all images of a female subjects belong to $f_1$, then the $\optdistnotation$ will assign $1/2$ probability to $f_1$ and $1/6$ to the rest of the $3$ female subjects.


\smallskip\noindent\textbf{Existence of Optimal Solution}
We now show the \emph{existence} of an optimal solution to our problem in~\eqref{eq:obj_constraint}.
Let $\region{y}{z} = \{ \x \colon f_y(\x) = y, f_z(\x) = z \}$ be a \emph{region} in the data space where the target attribute is $y$ and the sensitive attribute $z$.
Take $\mathcal{A}^*$ to be the following perturbation scheme: Any $\x\in \region{y}{z}$ is mapped to $\x'$ by (1) choosing a target region $\region{y}{z'}$ via probability $\optdist{y}{z'}$ over all $z' \in \mathcal{Z}$, and (2) choosing a datapoint $\x' \in \region{y}{z'}$ uniformly at random.
\begin{theorem}\label{theo:maximization}
$\mathcal{A}^*$ has the following properties:
\benum
\item $X^* = \mathcal{A}^*(X)$ is an optimal solution to \eqref{eq:obj_constraint}.
\item In presence of full prior knowledge about $P(z \vert y)$, the probability that any datapoint is observed in $X^*$ is equal to that of $X$.
\eenum
\end{theorem}
The proof is in Appendix~\ref{sec:theo_1_proof}.

\subsection{Privacy Guarantee}\label{sec:privacy_guarantee}

Now we show that the optimal solution, or an approximate solution with a small error, provides desirable privacy guarantees.
To model the correlations between $y$ and $z$, 
we use the Pufferfish framework~\cite{kifer2014puffer}, an extension of Differential Privacy (DP).

Let $\mathbb{S}$ be a set of secrets, $\mathbb{Q} \subseteq \mathbb{S} \times \mathbb{S}$ a set of secret pairs which we intend to make \emph{indistinguishable} and $\mathbb{D}$ the set of distributions governing the generation of data.
\begin{definition}\label{def:pufferfish}
 A mechanism $\mathcal{A}$ is said to be $(\varepsilon, \delta)$-Pufferfish($\mathbb{S}$,$\mathbb{Q}$,$\mathbb{D}$) private
 if for all $\x \in X$ drawn randomly from $\pi \in \mathbb{D}$, 
 for all secret pairs $(s_i, s_j) \in \mathbb{Q}$ for which $P(s_i \vert \pi)$ and $P(s_j \vert \pi)$ are not zero, and
 for all perturbed data $w \in \text{Range}(\mathcal{A})$
 we have:
\begin{equation}\label{eq:puff_original}
    P\left(\mathcal{A}(\x) = w \vert s_i, \pi\right) \leq
    e^\varepsilon P\left(\mathcal{A}(\x) = w \vert s_j, \pi\right) + \delta.
\end{equation}
\end{definition}
For brevity, we denote $(\varepsilon, \delta)$-Pufferfish($\mathbb{S}$,$\mathbb{Q}$,$\mathbb{D}$) by $(\varepsilon, \delta)$-Pufferfish.
The essential difference between DP and the Pufferfish framework is that the latter takes into account possible beliefs that an attacker might hold about the distribution of the data.
This is represented in our case by the prior knowledge set $\Theta$ which consists of (partial) distributions of sensitive attributes, $z$, conditioned on the target attributes, $y$. Note that this framework promises protection only to the extent that the attacker's knowledge on the relation between $y$ and $z$
allows.

We now proceed by defining $\mathbb{S}$, $\mathbb{Q}$ and $\mathbb{D}$ in our case.
Let $\mathbb{S}$ be the set of all possible values each $z_i$ can take given the corresponding $y_i$: $\mathbb{S} = \{\langle z_i = z \rangle \colon z \in \mathcal{Z}, \theta_{y_i}(z) \neq 0\}$,
$\mathbb{Q}$ be the set of all pairs of $z, z' \in \mathcal{Z}$ values for each $z_i$ such that an attacker cannot rule out given a target label $y_i$, i.e: either $\theta_{y_i}(z)$ does not exist or $\theta_{y_i}(z) \neq 0$ (similar for $z'$), and $\mathbb{D}$ be distributions of data given the knowledge about the target attribute for each data point, $y_i$.
\begin{theorem}\label{theo:pufferfish_guarantee}
Suppose that $\mathcal{A}$ is a randomized mechanism that produces perturbed data points
such that the predicted probabilities of $z$ follows a target distribution $\pi(z|y)$ within a bounded error $\gamma$:
\begin{equation}\label{eq:a_error}
\gamma = \sup_{z \in \mathcal{Z}, \x_i \in X} \big| \prob{f_z(\mathcal{A}(\x_i)) = z \mid y_i} - \pi(z|y_i) \big|,
\end{equation}
where $y_i$ is the corresponding target attribute for each $\x_i \in X$. Then,
$\mathcal{A}$ achieves $(0, 2\gamma)$-Pufferfish privacy.
\end{theorem}
The proof comes as a direct consequence of \eqref{eq:a_error}: for any $\x_i, \x_j \in X$ with the same target attribute ($y_i = y_j = y$), and $z \in \mathcal{Z}$ where $\theta_{y}(z) \neq 0$ we have:
\begin{equation}\label{eq:puff_bound}
\big| \prob{f_z(\mathcal{A}(\x_i)) = z \mid y} -  \prob{f_z(\mathcal{A}(\x_j)) = z \mid y} \big| \leq 2\gamma,
\end{equation}
which satisfies a $(0, 2\gamma)$-Pufferfish privacy. Note that $\pi(z|y)$ in Theorem~\ref{theo:pufferfish_guarantee} could be any arbitrary distribution, which entails that if one chooses $\optdistnotation$ as $\pi(z|y)$, Pufferfish guarantees are still provided. As a reminder, there are other reasons to choose $\optdistnotation$ that we reiterate here: following it ensures that (1) the performance of the target classifier does not suffer and (2) maximum \emph{uncertainty} is introduced for an adversary as an added measure of privacy protection.
\begin{remark}
$\mathcal{A}^*$ from Theroem~\ref{theo:maximization} achieves $(0, 0)$-Pufferfish privacy since $\gamma = 0$.
\end{remark}

Of course, in real-world scenarios, one rarely has a perfect predictor for $z$. Instead, both those who protect privacy and the attackers have \emph{approximations} of $f_z$. In Appendix~\ref{extention_approximate}, we show that if we use approximations of $f_z$ the error is factored into the Pufferfish privacy bound. 

\section{PR-GAN Design}\label{sec:design}
\def\bosch{OccuTherm}
We solve the optimization problem in \eqref{eq:obj_constraint} by using a modified GAN (Generative Adversarial Network) architecture. This section is devoted to design and implementation details.
\begin{figure}[tbh!]
    \centering
    \vspace*{-4mm}
        \includegraphics[width=0.34\textwidth]{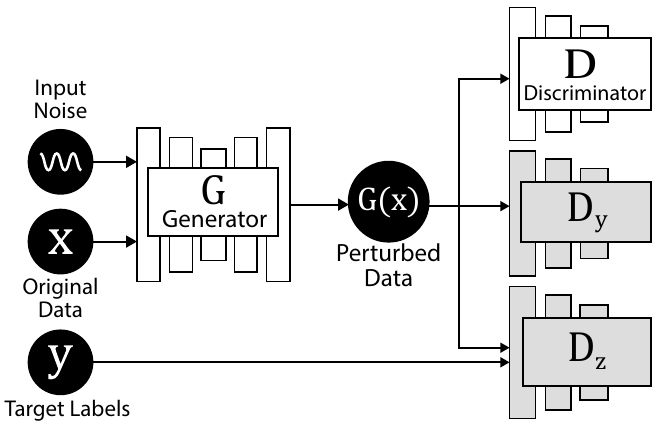}
    \caption{Architecture of proposed model. Target classifier $\DY$ and sensitive classifier $\DZ$ (in gray) are approximations of $f_y$ and $f_z$, pre-trained and fixed during training. $\G$ and $\D$ (in white) are trained in an adversarial setting.}
    \label{fig:architecture}
    \vspace*{-4mm}
\end{figure}
\subsection{Architecture}
In our architecture,we use two pre-trained classifiers $\DY$ and $\DZ$,
denoted by \emph{target} and \emph{sensitive} classifier, as approximations of $f_z$ and $f_y$. We also use a GAN, with a generative model $\G$ in charge of producing the perturbed data $\x' = \G(\x)$, and a discriminative model $\D$ which distinguishes $\x'$ from $\x$. $\mathcal{P}_\text{data}$ and $\mathcal{P}_g$ represent the distribution of original and perturbed data respectively.
Both $\DY$ and $\DZ$ are pre-trained classifiers plugged into our network. One can easily extend our model to employ multiple classifiers in cases where there are more than one sensitive or target attributes.
The overall structure of the network is shown in Figure~\ref{fig:architecture}. 

The process starts by generator $\G$ taking the original instance $\x$ as input and generating the perturbed version of the data, $\G(\x)$. Then $\G(x)$ is fed to the discriminators $\D$, $\DY$ and $\DZ$. $\D$'s goal is distinguishing real data from perturbed data. $\D(\x)$ represents the probability that $\x$ is an original datapoint ($\x \sim \mathcal{P}_{\text{data}}$) rather than one produced by $\G$ ($\x \sim \mathcal{P}_g$). We can write its loss function as below:
\begin{multline}\label{eq:d_loss}
\mathcal{L_{D}} = [\mathbb{E}_{\x \sim \mathcal{P}_{\text{data}}(\x)} \log \mathcal{D}(\G(\x)) \: +  \\ \mathbb{E}_{\x \sim \mathcal{P}_{\text{data}}(\x)} \log\big(1-\mathcal{D}(\x)\big)].
\end{multline}
$\G$ has multiple objectives. To ensure the utility of perturbed data while protecting privacy, we need to reintroduce our main objective function \eqref{eq:obj_constraint}. However, since we know that its second term, \eqref{eq:max_cond_entropy}, is maximized when the Maximum Entropy Distribution ($\optdistnotation$) is most faithfully followed, we can rewrite \eqref{eq:obj_constraint} as:
\begin{multline}\label{eq:adv_loss}
\mathcal{L_{\text{Adv}}} = \mathbb{E}_{\x} \big[ \ell(\DY(\G(\x), y) \: + \\
\crossentnotation\big(\DZ(\G(\x) \vert y), \optdist{y}{z}\big)
\big],
\end{multline}
where 
$\ell$ is a suitable loss function
and $\crossent{a}{b}$ the cross entropy between two distributions $a$ and $b$ which serves as a measure of their distance. Since we like the predicted probabilities for $z$, $\DZ(\G(\x) \vert y)$, to be as close as possible to $\optdist{y}{z}$ we minimize this distance.
$\G$, the GAN generator, wants to fool the GAN discriminator $\D$ in order to create perturbed data indistinguishable from real ones; the loss function for this will be:
\begin{equation}\label{eq:gan_loss}
\mathcal{L}_{\text{GAN}} = \mathbb{E}_{\x} \log{\big(1 - \D(\G(\x))\big)}.
\end{equation}
Finally, it has been shown that using regularization can greatly stabilize GAN's training and also provide an additional lever to control the utility of the perturbed data by limiting the overall perturbation one is allowed to add to the data~\cite{isola2017image}. Here, we use a \emph{hinge} loss:
\begin{equation}\label{eq:hinge_loss}
\mathcal{L}_{\text{hinge}} =
\mathbb{E}_{\x} \max\big(0, ||\G(\x) - \x|| - c\big),
\end{equation}
where $c$ is the maximum distance allowed before any loss occurs.

Our full objective for $\G$, combining \eqref{eq:adv_loss}, \eqref{eq:gan_loss} and \eqref{eq:hinge_loss}, is:
\begin{equation}
\mathcal{L}_{\G} = \mathcal{L}_{\text{GAN}} + \alpha\mathcal{L}_{\text{Adv}} + \beta\mathcal{L}_{\text{hinge}},
\end{equation}
where $\alpha$ and $\beta$ control the relative importance of the three losses. At each iteration, we alternate between training $\D$ and $\G$ while $\DZ$ and $\DY$ are previously trained and fixed in the network.

\subsection{PR-GAN Optimality}

In the original GAN design, the generator $\G$ defines a probability distribution $\mathcal{P}_g$ as the distribution of samples generated by $\G$. Under favorable assumptions, the distribution $\mathcal{P}_{g}$ converges to a good estimator of $\mathcal{P}_{\data}$ -- the distribution of training data~\cite{goodfellow2014generative}. In our case, as we introduce additional classifiers and gradients, we wish to understand how these classifiers, $\DZ$, $\DY$, modify the optimal solution and the final distribution. 
We follow the same assumption as in~\cite{goodfellow2014generative}: there is enough capacity and training time and the discriminator is allowed to reach its optimal given a fixed generator $\G$.

\paragraph{Fix $\G$, Optimize $\D$.} Notice that the discriminator $\D$ in our design uses the \emph{same} loss function as in the original GAN. So the following claim is still true.
\begin{lemma}[Optimal Discriminator~\cite{goodfellow2014generative}]
For a fixed generator $\G$, the optimal discriminator $\D$ is $$\D^*_{\G}(\x)=\frac{\mathcal{P}_{\data}(\x)}{\mathcal{P}_{\data}(\x)+\mathcal{P}_g(\x)},$$
where $\mathcal{P}_{\data}(\x)$ is the probability that $\x$ is from the data and $\mathcal{P}_g(\x)$ is the probability that $\x$ is from the generator $\G$.
\end{lemma}

\paragraph{Fix $\D$, Optimize $\G$.}
In the original GAN, the global minimum for the generator is achieved if and only if the generated distribution $\mathcal{P}_g$ is the same as $\mathcal{P}_{\data}$: $\mathcal{P}_g(x)=\mathcal{P}_{\data}(x)$.
Recall from Section~\ref{sec:definition} that this is exactly the second claim of Theorem~\ref{theo:maximization} when full knowledge about the conditional distribution $P(z \vert y)$ is allowed. This means that in this case, the mechanism $\mathcal{A}^*$ which achieves optimal results for \eqref{eq:obj_constraint} is also optimal as the GAN's generator $\G$.
In the absence of such knowledge, there will be tradeoff between the optimality of GAN's generator and causing maximum uncertainty about $P(z \vert y)$ for an attacker. In such a case, a suitable method would be to set a fixed performance threshold for $\DY$, the target classifier, and allow data to be perturbed only to the extent that $\DY$ performs above that threshold. We apply this technique in our experiments in Section~\ref{sec:experiment}.

\subsection{Implementation Details}\label{sec:implementation}
To minimize dependencies between different components in our model, we slice a dataset into 3 parts equal in size and class proportions. We use the first slice to train $\DY$ and $\DZ$, the second slice to train $\G$ and $\D$, and the third testing purposes.

We train and test the models on an Ubuntu $18.04.4$ machine with a GeForce GTX $1080$ Ti GPU. The code is written in Python using the TensorFlow $1.15.2$ Deep Learning framework. The training of PR-GAN consists of two steps: (1) the classifiers for the target and sensitive attributes, $\DY$ and $\DZ$ respectively, are trained, then (2) using those two, the noise generator, $\G$, is trained.
As we will see in Section~\ref{sec:mobile}, the noise-generating module $\G$ is light-weight and capable of perturbing data on embedded devices designed for ML applications in real time.

\section{Experiments}\label{sec:experiment}
\subsection{Datasets}\label{sec:dataset}
\def\wifi{WiFi}
\newcommand{\tr}{\text{train}}
\newcommand{\te}{\text{test}}
Below, we go over the $4$ datasets we have used along with sensitive and target attributes and the prior knowledge associated with each pair of attributes (recall that $z$ is the sensitive attribute while $y$ is the target attribute):
\begin{figure*}[!tbh]
    \centering
    \includegraphics[width=0.85\textwidth]{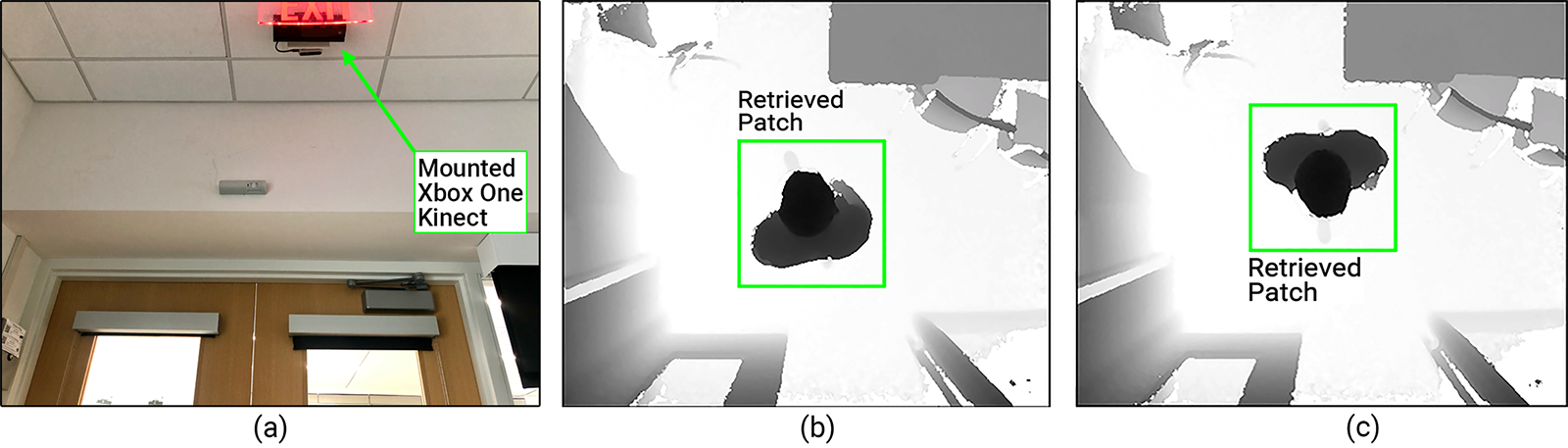}
    \caption{An Xbox One Kinect mounted above an entrance door (a) along with a sample depth frame of someone facing inside (b) and outside (c) in the \bosch~ dataset. The square patches in (b) and (c) are retrieved from images to use for recognition.} \vspace*{-4mm}
    \label{fig:in_and_out}
\end{figure*}

\smallskip\noindent\textbf{MNIST}~\cite{MNISTDataset}: A dataset of $70{,}000$ handwritten letters. For this dataset, $y$ is the parity (being odd or even) of the numbers and $z$ is whether or not a digit is less than $5$; a toy example. The prior knowledge, $\theta_E$ for \emph{evens} and $\theta_O$ for \emph{odds}, is as follows:
    \begin{equation*}\label{eq:mnist_theta}
    \theta_O(z) =
    \begin{cases}
    0.4&  \text{if } z < 5\\
    0.6&  \text{if } z \geq 5
    \end{cases}, \;\;
    \theta_E(z) =
    \begin{cases}
    0.6&  \text{if } z < 5\\
    0.4&  \text{if } z \geq 5
    \end{cases}.
    \end{equation*}
\smallskip\noindent\textbf{PubFig Faces}~\cite{FaceDataset}: This dataset includes 58,797 images of 200 people. 
We define $z$ as the identity of each subject while $y$ is each person's gender, which can happen in a scenario where subjects are donating their images to train a gender classifier.
We use MTCNN~\cite{MTCNN} to align the faces and remove duplicates for each person. Filtering out subjects with less than $200$ images, we will have $5$ female and $10$ male subjects with a total of $6{,}553$ images.
For the prior knowledge, $\theta_M$ for men and $\theta_F$ for women, we have:
    \begin{equation*}\label{eq:pubfig_theta}
    \theta_M(z) =
    \begin{cases}
    1/10&   z \in \mathcal{Z}_m\\
    0&  z \in \mathcal{Z}_f
    \end{cases}, \:
    \theta_F(z) =
    \begin{cases}
    0&  z \in \mathcal{Z}_m\\
    1/5&  z \in \mathcal{Z}_f
    \end{cases},
    \end{equation*}
    where $\mathcal{Z}_m$ and $\mathcal{Z}_f$ are the set of male and female subjects.

\smallskip\noindent \textbf{UJI Indoor Localization (\wifi)}~\cite{WiFiDataset}: In this dataset, signal strengths of $520$ WiFi access points (WAP) are recorded for $21{,}048$ locations inside $3$ different buildings. The buildings have a total of $13$ floors. In addition, each data record is associated with one of $8$ regions in each floor.
Similar to the scenario discussed in Section~\ref{sec:intro},
we define $z$ as the specific region a user is in on each floor (one of $104$ total possibilities), and $y$ as the floor on which the user is at any moment (one of $13$ options).
The strength signals, originally continuous, are turned into binary attributes that represent the ``existence'' of any signal from a WAP; this proved to produce the best performance in our experiments.
For each floor $y \in [13]$, there is set of \emph{valid} regions, $\mathcal{Z}_y$, which includes the $8$ regions associated with that floor. For $\theta_y(z)$ we have:
\begin{equation*}
    \theta_y(z) = \begin{cases}
    0&  z \notin \mathcal{Z}_y\\
    1/8&  z \in \mathcal{Z}_y
    \end{cases}.
\end{equation*}
    
\smallskip\noindent\textbf{\bosch}~\cite{OccuTherm_dataset}: This dataset consists of depth frames taken by a Kinect for XBOX One mounted above the entrance door of a conference room\cite{OccuTherm_dataset}. It includes frames of $77$ subjects facing inside the room, and then facing outside of it.
The installation setting, along with a sample of the resulting two images can be found in Figure~\ref{fig:in_and_out}. Each depth frame has a resolution of $512 \times 424$. From each frame, we retrieve a $180 \times 180$ patch centered around the participant in the frame. Here, we define $z$ as the identity of users, while $y$ is whether a subject is facing inside, or outside the room; a binary attribute.
To rely on subjects with enough images available, we limit the dataset to the top $20$ participants with the most number of images. That leaves us with $60{,}033$ images in total. Since $y$ and $z$ have no correlation, the prior knowledge for the conditional distribution between $z$ and $y$ is a uniform distribution:
\begin{equation*}
    \theta_y(z) = 1/n,
\end{equation*}
where $n$ is the total number of people in the dataset, $20$ here.
The architecture for the neural networks $\DY$ and $\DZ$ used for each dataset is available in Table~\ref{tab:architecture} (``VGG16 base'' refers to the bottom layers of VGG16~\cite{VGG16} architecture with modified top layers).
\begin{table}[bth]
    \centering
    \caption{Model architectures for the $4$ datasets  (Conv: convolution layer, FC: fully-connected layer).}
    \begin{small}
    \begin{tabular}{cccc}
        \toprule
        MNIST & PubFig & \wifi & \bosch \\
        \midrule
        Conv(64,5,5)  & VGG16 base   & FC(256)      & VGG16 base    \\
ReLu          & FC(1024)     & ReLu         & FC(1024)      \\
Conv(64,5,5)  & ReLu         & Dropout(0.5) & ReLu          \\
ReLu          & Dropout(0.5) & FC(128)      & Dropout(0.5)  \\
Dropout(0.25) & FC(512)      & ReLu         & FC(512)       \\
FC(128)       & ReLu         & Dropout(0.5) & ReLu          \\
ReLu          & FC(Dim(Y))   & FC(64)       & FC(Dim(Y))    \\
Dropout(0.5)  &              & ReLu         &               \\
FC(2)         &              & FC(Dim(Y))   &               \\
        \bottomrule
    \end{tabular}
    \end{small}
\label{tab:architecture}
\end{table}
\subsection{Baseline}
\def\Lap{\operatorname{Lap}}
\def\Dim{\operatorname{Dim}}
In our experiments, we compare our method against the following baselines:
\begin{figure}
    \centering
    \includegraphics[width=\columnwidth]{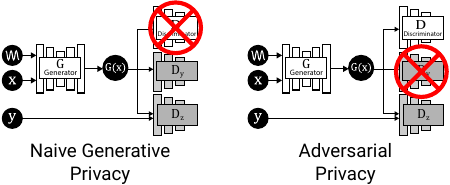}
    \caption{Two baseline approaches created by removing (left) the GAN's discriminator, $\D$, and (right) the target classifier.
    } \vspace*{-4mm}
    \label{fig:baseline}
\end{figure}
\begin{table*}[!tbh]
    \centering
    \caption{Sensitive classifier accuracy, $Acc(\DZ)$, across all methods over $4$ datasets. Lower is better.
    }\vspace*{-4mm}
    \label{tab:performance}
    
    \begin{threeparttable}
    \begin{tabular}{lcccccccc}
    \toprule
    \multirow{3}[3]{*}{\large{Method}} & \hspace{.2cm} &
    \multicolumn{7}{c}{Sensitive Classifier Accuracy (\%)} \\
    \cmidrule(lr){3-9}
    & &
    MNIST & \hspace{.2cm} &
    PubFig & \hspace{.2cm} &
    \wifi & \hspace{.2cm} &
    \bosch \\
    & &
    \small{$Acc(\DY) \geq 0.95$} & &
    \small{$Acc(\DY) \geq 0.95$} & &
    \small{$Acc(\DY) \geq 0.75$} & &
    \small{$Acc(\DY) \geq 0.95$} \\
    \midrule
    PR-GAN & &
    \textbf{65.3} & &
    \textbf{20.6} & &
    \textbf{23.6} & &
    \textbf{3.6} \\
    NGP & &
    67.2 & &
    28.1 & &
    31.3 & &
    7.2 \\
    AP & &
    67.1 & &
    61.9 & &
    37.7 & &
    23.6 \\ 
    DP & &
    80.6 & &
    78.3 & &
    50.0 & &
    37.6 \\
    \midrule
    Random\tnote{\textasteriskcentered} & &
    60.0 & &
    13.3 & &
    12.5 & &
    5.0 \\
    Original $Acc(\DZ$) & &
    98.4 & &
    80.7 & &
    84.6 & &
    99.8 \\
    Original $Acc(\DY)$ & &
    99.2 & &
    98.3 & &
    92.6 & &
    99.95 \\
    \bottomrule
    \end{tabular}
    \begin{tablenotes}
     \footnotesize
     \item[\textasteriskcentered] A classifier that outputs a class at random according to $\theta_y(z)$.
    \end{tablenotes}
    \end{threeparttable} \vspace*{-4mm}
\end{table*}

\textbf{Naive Generative Privacy (NGP)}:
Some have previously proposed the use of $\DY$ and $\DZ$, trained in competition, without the GAN's discriminator, $\D$. To show that this module contributes positively to the performance of $\G$, we test our results against a baseline where $\D$ is removed, calling it \emph{Naive Generative Privacy} (NGP). The architecture is depicted 
on Figure~\ref{fig:baseline}, left.

\textbf{Adversarial Privacy (AP)}:
Another line of work similar to ours proposes the production of adversarial samples against $\DZ$ with a limit on the amount of noise, similar to our regularization at \eqref{eq:hinge_loss}, to ensure high utility. These methods lack the presence of $\DY$ to directly guide the perturbations towards higher utility. To test the effect of $\DY$'s absence, we remove this module, keeping the rest of the architecture similar, as seen on Figure~\ref{fig:baseline}, right, calling it \emph{Adversarial Privacy} (AP).

\textbf{Differential Privacy (DP)}:
To highlight the fact that guidance on both $\DY$ and $\DZ$ is essential, we show that conventional methods, such as DP fail to achieve a desirable trade-off between utility and privacy.
For real-valued vectors (image datasets), one can use the \emph{Laplace Mechanism} known to achieve $\varepsilon$-differential privacy, where independent noise is added to each pixel via the Laplacian distribution:
\begin{equation*}
\Lap(z \vert b) = \frac{|z|}{2b}\exp{\Big(-\frac{1}{b}\Big)},
\end{equation*}
where $b$ is the scale parameter. This method achieves $1/b$-differential privacy~\cite{Dwork2014-pq}.

Not all of the baseline approaches have a mechanism to assure high utility. We make results comparable by setting a threshold for $\DY$'s accuracy score (denoted by $Acc(\DY)$) and add as much noise as possible without going below that threshold when $\DY$ is tested on resulting perturbed data.
Our method is denoted by PR-GAN throughout experiments.
\subsection{Running on Mobile Devices}\label{sec:mobile}
As discussed in Section~\ref{section:system}, once trained, the goal is to install the trained generator ($\G$) on remote sensing devices to produce perturbations from source. This has the advantage that \emph{non-perturbed} will never leave local devices.
The complexity and efficiency of a neural network depends on many factors. However, as is common practice, we measure it by counting the number of parameters in a network and the number of floating-point operations (FLOP). In  Table~\ref{tab:complexity}, we compare the complexity of our networks with state-of-the-art networks designed specifically to run on embedded devices. As you can see, PR-GAN is more compact and computationally cheaper compared to the state-of-the-art across all $4$ datasets, making them suitable to run on current mobile devices.
In Table~\ref{tab:fps}, we have measured the running time and the corresponding frame-per-second (\emph{input}-per-second in the case of \wifi~dataset) capability of PR-GAN for the $4$ datasets.
Using the running time measured on a GeForce GTX $1080$ Ti GPU across the $4$ datasets, we use the latest benchmarks for $3$ devices designed for ML applications, namely Nvidia's Jetson Nano, Raspberry Pi 3 with an Intel NCS2, and Coral Dev Board, to estimate the running time of PR-GAN on these devices~\cite{rpi-benchmark,jetson-benchmark} using linear regression.
The two smaller datasets, \wifi~ and MNIST, allows for high-throughput generation of noise by our generator model, up to hundreds per second and easily handling any scenario. For the two more complicated datasets, PubFig and \bosch, with an acceptable rate to allow PR-GAN to be used in real-life settings; in the case of \bosch~dataset, we achieve more than $10$ fps while $3$ fps is known to be good enough for real-time tasks~\cite{FORK}.
Although the estimated values might have inaccuracies, the resulting values are, in virtually all of the cases, at least one order of magnitude greater than what is required.
\begin{table}[!htb]
\centering
\caption{The complexity of our models compared to state-of-the-art models designed for ImageNet~\cite{krizhevsky2012imagenet} classification task on mobile devices.}
\label{tab:complexity}
\begin{threeparttable}
\vspace*{-4mm}
\begin{small}
\begin{tabular}{lcc}
\toprule
Network & FLOP & Parameters\\
\midrule
MobileNetV1 1.0~\cite{MobileNetV1}     & 575M & 4.2M   \\
ShuffleNet 1.5x~\cite{ShuffleNet} & 292M & 3.4M   \\
NasNet-A~\cite{NasNet}  & 564M & 5.3M   \\
MobileNetV2 1.0~\cite{MobileNetV2}    & 300M & 3.4M   \\
\midrule
\multicolumn{3}{c}{Our Noise Generators}\\
\midrule
MNIST           & 1.6M   & 235.4K \\
PubFig          & 232M  & 644.2K \\
\wifi            & 2.1M   & 1.1M \\
\bosch            & 96.4M   & 420.3K \\
\bottomrule
\end{tabular}\end{small}
\vspace*{-4mm}
\end{threeparttable}
\end{table}

\begin{table}[!htb]
\centering
\caption{PR-GAN noise generation speed, measured by frames (\emph{input} in the case of \wifi~dataset) per second, for the $4$ datasets on $4$ devices.}
\label{tab:fps}\vspace*{-4mm}
\begin{threeparttable}
\begin{small}
\begin{tabular}{lccccc}
\toprule
\multirow{3}[2]{*}{Network} & 
 GTX & Jetson & RPI3 + & Coral  \\
&  $1080$ Ti & Nano\tnote{\textasteriskcentered} & Intel NCS2\tnote{\textasteriskcentered} & Dev Board\tnote{\textasteriskcentered} \\
\midrule
MNIST & $22.1$K & $1.7$K & $826.4$	& $3.9$K \\
PubFig & $372.0$ & $29.1$ & $10.7$ & $42.2$ \\
\wifi & $36.0$K & $2.8$K & $1.4$K & $6.4$K \\
\bosch & $595.8$ & $46.7$ & $19.1$ & $82.0$ \\
\bottomrule
\end{tabular}
\end{small}
\begin{tablenotes}
 \footnotesize
 \item[\textasteriskcentered] Values are estimated based on running time on GeForce GTX $1080$ Ti GPU.
\end{tablenotes} \vspace*{-4mm}
\end{threeparttable}
\end{table}
\begin{figure*}
    \centering
    \includegraphics[width=0.95\textwidth]{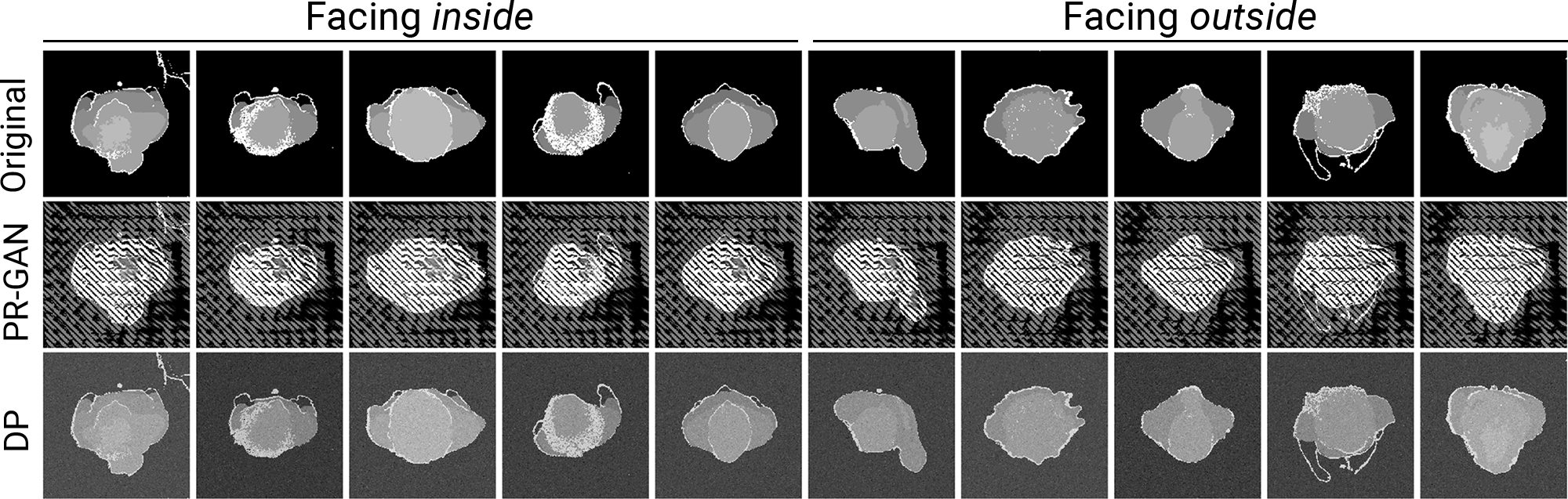}
    \caption{A randomly selected sample of images from the \bosch~dataset. The first row includes the original patches and the second and third rows include perturbed images using PR-GAN and Laplacian noise (DP) respectively. In the $5$ columns on the left the subject is facing \emph{inside} while in the $5$ columns on the right (s)he is facing \emph{outside}.} \vspace*{-4mm}
    \label{fig:bosch_sample}
\end{figure*}
\subsection{Performance}\label{sec:performance}
To show our method's ability to preserve $y$ while causing uncertainty about $z$,
we compare the resulting
classification accuracy for sensitive attributes, $Acc(\DZ$, with a fixed utility, $Acc(\DY)$, threshold.
For the the \wifi~dataset, we choose $75\%$ as the threshold for $Acc(\DY)$ and for the $3$ image datasets, namely MNIST, PubFig and \bosch, we choose $95\%$. These thresholds are selected based on the difficulty of achieving acceptable privacy guarantees and what an acceptable value for $Acc(\DY)$ is for each dataset.
For methods based on DP, the optimal value of DP's $\varepsilon$ is found by iterating over different values of $\varepsilon$ from $0.01$ to $10$ and selecting the smallest $\varepsilon$ (corresponding to the largest added noise) where $Acc(\DY)$ is still above the set threshold and reporting the resulting
$Acc(\DZ)$.

Results are available in Table~\ref{tab:performance}. The original performance of $\DY$ and $\DZ$ are available in the two rows at the bottom. The performance of a random classifier which uses the prior knowledge, $\theta_y(z)$, is reported in the third row from the bottom. Note that this value is theoretically the lowest possible accuracy a privacy mechanism can guarantee for the classification of $z$.

First, note that our method outperforms all others in all $4$ datasets. In the case of the \bosch~ dataset, PR-GAN produces the theoretically optimal perturbations; it decreases $Acc(\DZ)$ to its lowest possible value (the row denoted by ``Random*'' in Table~\ref{tab:performance}) while maintaining an accuracy of $99.6\%$ for $Acc(\DY)$, just $0.2\%$ below its original value, $99.8\%$. A randomly selected sample of $10$ individual patches, $5$ looking inside and $5$ looking outside, are depicted in Figure~\ref{fig:bosch_sample} along with the corresponding perturbed versions of each image produced by PR-GAN and DP.
Next, notice that the two methods that do not utilize the target attributes in producing perturbations (DP and AP) achieve results that are far less promising than the other two methods (PR-GAN and NGP).
The discrepancy is far more pronounced in the two image datasets, PubFig and \bosch, where producing perturbations is harder due to the complex nature of the data and the higher number of pixels in play.
Compared to NGP, our method consistently achieves better results. This reveals that the existence of the full structure of a GAN, and specifically the GAN discriminator $\D$, is effective in producing perturbations that preserve high utility for the data. We dive deeper into the difference between the two Section~\ref{sec:tradeoff}. Producing noises for the \wifi~ dataset proved to be the most difficult task, however, PR-GAN achieves considerably better performance than NGP in this case as well. There are two main reasons for this difficulty. First, the conventional neural networks for image data are far more advanced than those used on arbitrary feature vectors, such as the one in the \wifi~dataset. For instance, to produce gradients for discrete-valued feature vectors, one has to apply complicated techniques, none of which is necessary when one works with image data. Second, the target and sensitive attributes in this case are \emph{highly} correlated; for any sensitive attribute (a region on a floor), only $1$ out of $13$ possible target attributes are valid, which heavily restricts the space in which PR-GAN can search for optimal solutions.
\begin{figure}[!tbh]
    \centering \vspace*{-4mm}
    \includegraphics[width=0.8\columnwidth]{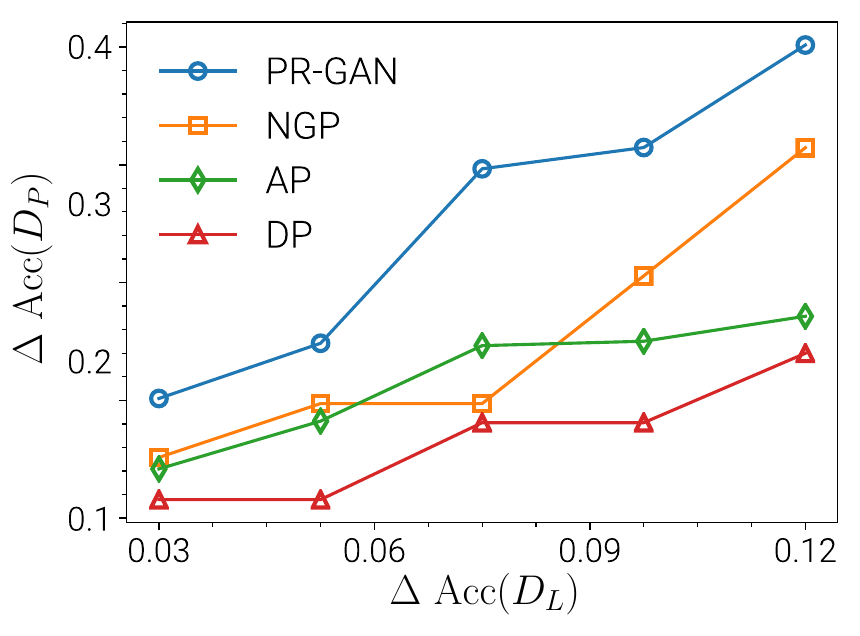} \vspace*{-2mm}
    \caption{The trade-off between achieved privacy ($\Delta~Acc(\DZ)$) and utility loss budget ($\Delta~Acc(\DY)$ for the \wifi~ dataset.}
    \label{fig:tradeoff}\vspace*{-4mm}
\end{figure}
\subsection{Utility vs. Privacy}\label{sec:tradeoff}
Inherent in any privacy protection scenario is the trade-off between privacy and utility. To show our method's superior ability to find a good trade-off, we conduct an experiment on the \wifi~dataset in which different \emph{utility budgets}, in terms of how low can $Acc(\DY)$ get, is given to the $4$ methods. 
The smaller size of the \wifi~dataset is what allows us to do such an exhaustive experiment; doing so on the other $3$ datasets was not possible given resources available to us.
The budget, $\Delta Acc(\DY)$, is set to $5$ values between $0.03$ and $0.12$.
The results are depicted in Figure~\ref{fig:tradeoff}. As you can see, our method is consistently above the other $3$, and the two methods that do not utilize the target labels in any way, namely AP and DP, have a significantly slower gain in privacy guarantee (the amount $Acc(\DZ)$ is decreased) as the utility budget increases.
\subsection{Transferability}\label{sec:transfer}
In an identity-masking scenario such as the one in our experiment with the \bosch~dataset, it might not be possible to have every occupant of a building to contribute data to the training of PR-GAN. The occupants change over time,
the manufacturer of PR-GAN might be a different entity than the building managers and occupants, and most importantly, by collecting raw images of every occupant, we endanger their privacy in the first place.
To see if an entity, such as the manufacturer, can train a model on a group and deploy it on a completely different group, we conduct the following experiment:
We use the noise generator from Section~\ref{sec:performance}, trained on the top-$20$ individuals with most images, and produce perturbations for the next $30$ individuals with the highest number of images, each having at least $1000$ images. Then, using an identity-classifier trained exclusively on this new group of $30$ individuals, we test the resulting perturbed methods. We observe that $Acc(\DZ) = 3.71\%$ and $Acc(\DY) = 86.9\%$, showing minimal performance loss on $y$ and optimal privacy on $z$. This means that our model's ability to hide identity-revealing features transfers from one group to another, completely new group, not-seen-before during training. This is promising as it allows a manufacturer to train $\G$ in a warehouse and ship it to different sites for deployment, expecting high performance.
\subsection{Training Utility}\label{sec:new_training}
\begin{table}[!htb]
\centering \vspace*{-4mm}
\caption{Utility of the perturbed data when used for training a new model, compared to when it is used for inference (Table~\ref{tab:performance}).}\vspace*{-4mm}
\begin{small}
\begin{tabular}{lccccc}
\toprule
\multirow{2}[3]{*}{Application} & \hspace{0.1cm} & \multicolumn{4}{c}{Target Classifier Accuracy (\%)}
\\
\cmidrule(lr){3-6}
& & MNIST & PubFig & \wifi & \bosch \\
\midrule
Inference & & 95.19 & 95.47 & 79.80 & 99.82 \\
Training & & 96.72 & 98.84 & 72.06 & 99.73 \\
\bottomrule
\end{tabular} \vspace*{-4mm}
\label{tab:new_training}
\end{small}
\end{table}
We finally test whether the data perturbed by PR-GAN could be used to train a new model, which can perform well on raw, clean data. This is Inspired by a scenario where a model is trained by public, perturbed data, and is then deployed to personal devices to perform classification on raw, clean data.
The results are available in Table~\ref{tab:new_training}. The first row is a refresher on Section~\ref{sec:performance}, where a model was trained on clean data and inference was done on perturbed data (the opposite of what we do in this section). The second row includes the accuracy of $\DY$ when perturbed data is used for training and clean data (from a separate slice) is used for testing. As you can see, overall, the performance values are close in both cases across all datasets.

\section{Conclusion and Future Work}\label{sec:conclusion}

In this work, we designed and implemented a framework to bridge the gap between privacy preserving data publishing and deep generative models.
We showed that it is possible to use deep neural networks as clues for generating tailored perturbations, which successfully hides sensitive information while offering a high degree of utility for both inference and training tasks. By choosing this approach, not only we can effectively protect sensitive information, but we can also maintain the information necessary for a given target application. Future work includes finding specialized architectures to perturb different types of data (e.g.: time series) and improving on the privacy guarantees we provide.

\noindent\textbf{Acknowledgement:} The authors would like to acknowledge support in part from NSF CNS-1618391, DMS-1737812, OAC-1939459, DOE DE-EE0007682. The opinions expressed here are those of the authors and do not necessarily reflect the views of the funding agencies.

\balance
\bibliographystyle{abbrv}
\bibliography{main,privacy,private-sensing,sensor-system}

\appendix
\section{Proof of Theorem 1}\label{sec:theo_1_proof}
\begin{proof}
The first claim follows by definition. First $\mathcal{A}^*$ follows the $\optdistnotation$, and second the corresponding label $y$ for each entry is untouched, meaning $\prob{f_y(\x) = y} = \prob{f_y(\mathcal{A}^*(\x) = y}$. This shows that the loss value will remain optimal for $f_y$.

We now focus on the second claim. By looking at \eqref{eq:max_entropy_dist} one can realize that if for all $y \in \mathcal{Y}$ and $z \in \mathcal{Z}$, $\theta_y(z)$ exists, then $\optdist{y}{z}$ becomes the real distribution of $z$ conditioned on $y$: $P(z \vert y)$. This means that $\prob{\mathcal{A}^*(\x) \in \region{y}{z}} = P(z \vert y)$. Knowing this, we can write about the probability of observing $\x'$ in a region $\region{y}{z}$ for any $z$ as:
\begin{align*}
    \prob{\x' \in \region{y}{z}} {}&=
    \sum_{z' \in \mathcal{Z}}\prob{\x \in \region{y}{z'}}P(z \vert y)\\
    &= P(z \vert y) \sum_{z'}\prob{\x \in \region{y}{z'}} \\
    &= P(z \vert y) \Pr_{\x \sim X}\left[ f_y(\x) = y \right] \\
    &= \Pr_{\x \sim X}\left[\x \in \region{y}{z'}\right].
\end{align*}
which concludes the proof.
\end{proof}
\section{Extension to Approximate Predictors}\label{extention_approximate}

Suppose that $g_z$ and $h_z$ are two approximations of $f_z$, the former used by the protector of privacy to produce perturbations and the latter by an attacker to reveal sensitive information. Suppose also that $h_z$ and $g_z$ have a bounded approximation error $\epsilon$ such that for all $\x \in X$:
\begin{equation}\label{eq:classifier_error}
\begin{aligned}
    \left|\prob{g_z(\x) = z \vert y} - \prob{f_z(\x) = z \vert y}\right| &\leq \epsilon \\
    \left|\prob{h_z(\x) = z \vert y} - \prob{f_z(\x) = z \vert y}\right| &\leq \epsilon.
\end{aligned}
\end{equation}
\begin{theorem}\label{theo:pufferfish_extension}
Any privacy mechanism $\mathcal{A}$ that is $(\varepsilon, \delta)$-Pufferfish for $g_z$ is $(\varepsilon, \delta + 2\epsilon(e^\varepsilon + 1))$-Pufferfish for $h_z$.
\end{theorem}
\begin{proof}
Given the conditions in \eqref{eq:classifier_error}, we can write:
\begin{equation}\label{eq:classifier_distance}
\left|\prob{g_z(\x) = z \vert y} - \prob{h_z(\x) = z \vert y}\right| \leq 2\epsilon.
\end{equation}
By $(\varepsilon, \delta)$-Pufferfish guarantees we have:
\begin{equation}\label{eq:puff_guarantee}
    \prob{g_z(\x) = z_1 \vert y} \leq e^\varepsilon \prob{ g_z(\x) = z_2 \vert y} + \delta.
\end{equation}
Incorporating \eqref{eq:classifier_distance} in \eqref{eq:puff_guarantee} yields:
\begin{gather}
    \prob{h_z(\x) = z \vert y} - 2\epsilon \leq e^\varepsilon \left(\prob{h_z(\x) = z \vert y} + 2\epsilon \right) + \delta \nonumber\\
    \prob{h_z(\x) = z \vert y} \leq e^\varepsilon \prob{ h_z(\x) = z \vert y} + 2\epsilon(e^\varepsilon + 1) \delta, \nonumber
\end{gather}
which concludes the proof.
\end{proof}
Theorems~\ref{theo:pufferfish_guarantee} and \ref{theo:pufferfish_extension} result in the following.
\begin{corollary}
Against an attacker who uses an unseen predictor of $z$ with bounded error $\epsilon$, we guarantee $(0, 2\gamma + 4\epsilon)$-Pufferfish privacy.
\end{corollary}

\end{document}